\documentclass{article}

\PassOptionsToPackage{numbers, compress}{natbib}
\usepackage[preprint]{neurips_2026}

\usepackage[utf8]{inputenc} %
\usepackage[T1]{fontenc}    %
\usepackage{hyperref}       %
\usepackage{url}            %
\usepackage{booktabs}       %
\usepackage{amsfonts}       %
\usepackage{amsmath}
\usepackage{amssymb}
\usepackage{amsthm}
\usepackage{nicefrac}       %
\usepackage{microtype}      %
\usepackage{xcolor}         %
\usepackage{graphicx}
\usepackage{caption}        %
\usepackage{float}          %
\usepackage{enumitem}
\usepackage{algorithm}
\usepackage{algpseudocode}

\newtheorem{theorem}{Theorem}

\newcommand{\R}{\mathbb{R}}
\newcommand{\E}{\mathbb{E}}

\newcommand{\Sd}{S^{d-1}}
\newcommand{\Phimap}{\Phi}
\newcommand{\unif}{\mathrm{Unif}}

\newcommand{\Lrep}{\mathcal{L}_{\mathrm{rep}}}
\newcommand{\Lrad}{\mathcal{L}_{\mathrm{rad}}}
\newcommand{\Lmom}{\mathcal{L}_{\mathrm{mom}}}
\newcommand{\Lwb}{\mathcal{L}_{\mathrm{wb}}}
\newcommand{\indep}{\perp\!\!\!\perp}

\title{The Wristband Gaussian Loss: Deterministic, Composable Latents via a Sphere--Interval Decomposition}

\author{%
  Mikhail Parakhin \\ %
  Shopify\\
  \texttt{mikhail.parakhin@shopify.com} \\
  \And
  André M. Carvalho \\
  \texttt{ac@andremiguel.pt} \\
  \And
  Patrick Haluptzok \\
  Shopify \\
  \texttt{patrick.haluptzok@shopify.com} \\
}

\begin{document}

\maketitle

\begin{abstract}
We present the \emph{Wristband Gaussian Loss}, a deterministic batch
loss for Gaussianizing point embeddings without sampling, KL terms,
or iterative transport. Each $x \in \mathbb{R}^d$ is mapped to a
direction $u = x/\|x\|$ and a CDF-transformed radius
$t = F_{\chi^2_d}(\|x\|^2)$ on the wristband $S^{d-1} \times [0,1]$.
We prove (and machine-verify in Lean~4) that for $d \ge 2$ the
pushforward wristband map equals
$\sigma_{d-1} \otimes \mathrm{Unif}[0,1]$ iff the source is
$\mathcal{N}(0, I_d)$, and that the Neumann-reflected wristband
repulsion energy is uniquely minimized at the uniform target. We
compute this reflected-kernel objective in two ways: a nearest
three-image pairwise truncation at $O(N^2 d)$, and a spectral
Neumann path joining angular and radial Mercer modes
(spherical-harmonic and cosine) at $O(N d K)$, with empirically
matched gradients. A 1D Wasserstein radial term and a moment
penalty serve as finite-sample accelerators with the same optimum,
and Monte-Carlo null calibration turns the components into a single
standardized statistic. We evaluate direct point-cloud
Gaussianization with a calibrated barycentric $W_2$ score: a
deterministic Gaussian reference batch is built by recursive Hungarian
averaging, with each method reported as a $z$-score against
same-size Gaussian batches. On the axis-uniform X benchmark,
Wristband is competitive in 2D and gives the best 10D score. On a harder radial--angular-copula impostor whose
Gaussian radial and angular marginals are correct but dependent,
Wristband gives the best 10D and 128D scores. Coupled with learnable-key Euclidean attention and
exact invertible flows, the resulting Deterministic Gaussian
Autoencoder delivers a Gaussian-latent interface for counterfactual
sampling with independent factors and a context/residual
construction for dependent factors.
\end{abstract}

\section{Introduction}

A long-standing goal of representation learning is a
\emph{deterministic} encoder
$E: \R^{d_{\text{in}}} \to \R^d$ whose output matches a fixed prior,
typically $\mathcal{N}(0, I_d)$. A Gaussian latent code is desirable
for three reasons: (i)~the standard normal density factorizes across
coordinates, so disjoint blocks of dimensions are independent;
(ii)~the prior is closed under marginalization and conditioning,
enabling counterfactual sampling and submodule swap; and
(iii)~Gaussian latents are the maximum-entropy distribution at fixed
second moment, without explicit InfoMax estimators. None hold for
popular alternatives: VAE-style stochastic encoders inject sampling
noise on every forward pass;
sphere-uniformity losses~\citep{wang2020understanding} couple
coordinates through the norm; diffusion-based
pipelines~\citep{ho2020denoising,lipman2023flow} require iterative
inference.

The technical obstacle is that, in the deterministic setting, one
observes only a finite batch of point embeddings rather than a
per-sample posterior. A useful loss must therefore answer ``does
this batch of $N$ points look like $N$ i.i.d.\ draws from
$\mathcal{N}(0, I)$?'' under three constraints: (a)~strong enough
to penalize \emph{higher-order} dependencies missed by moment-based
methods (multimodality, heavy tails); (b)~stable and well-scaled
across moderate-to-high dimensions $d$; and (c)~GPU-friendly ---
ideally $O(N^2)$ or better, without requiring an inner optimal-transport solve.

Existing approaches each compromise on one of these axes. Moment
matching constrains only the first two moments;
MMD~\citep{gretton2012kernel} depends on bandwidth and dimension;
population sliced Wasserstein distances are characteristic by
Cram\'er--Wold, but finite-projection estimators have weak signal on
localized joint clumping and ignore the radial--angular structure of
Gaussians~\citep{kolouri2019sliced}; Hungarian-matched optimal
transport is $O(N^3)$ and better suited to evaluation than training.
Recent radial extensions of variance-covariance
regularization~\citep{kuang2025radial} match the radial
\emph{marginal} to a Chi distribution but, by their own analysis,
are sufficient for Gaussianity only under elliptical symmetry --- a
condition that learned features rarely satisfy.

\paragraph{Contributions.} This paper introduces the \emph{Wristband Gaussian Loss} and the surrounding deterministic autoencoder framework. We make four contributions:
\begin{enumerate}[leftmargin=*]
  \item \textbf{A biconditional characterization of $\mathcal{N}(0, I_d)$ via a sphere--interval decomposition.} We define the wristband map $\Phimap(x) = (x/\|x\|, F_{\chi^2_d}(\|x\|^2))$, taking values in $\Sd \times [0, 1]$, and prove that uniformity on this product space is equivalent to $\mathcal{N}(0, I_d)$ for $d \ge 2$ (Theorem~\ref{thm:equivalence}). The result is machine-verified in Lean~4 (\S\ref{sec:formalization}).
  \item \textbf{A principled pairwise repulsion loss with reflected-boundary correction.} We construct the full reflected-Neumann wristband repulsion whose unique population minimizer is the wristband-uniform law; radial Wasserstein and moment matching are optional finite-sample accelerators with the same target (\S\ref{sec:loss}). The implemented pairwise loss uses the nearest three images: the real point, its reflection across $0$, and its reflection across $1$.
  \item \textbf{A spectral fast path.} We derive a Neumann--cosine $\times$ spherical-harmonic approximation that reduces the large-batch repulsion computation from $O(N^2 d)$ to $O(NdK)$ (\S\ref{sec:spectral}). The spectral path is explicitly an approximation to the pairwise reflected kernel, and we validate its gradient parity empirically.
  \item \textbf{Empirical validation.} On axis-uniform X-shaped synthetic distributions and on a harder radial--angular copula impostor, calibrated barycentric $W_2$ scores show the benefit of targeting the joint direction--radius law rather than only moments or radial marginals. Wristband is strongest in the higher-dimensional settings: it gives the best 10D score on X-shaped data and the best 10D and 128D scores on the copula benchmark. On a 15-dimensional non-Gaussian mixture and on MNIST conditional inpainting, the resulting Deterministic Gaussian Autoencoder turns this distributional target into an end-to-end deterministic latent interface (\S\ref{sec:experiments}).
\end{enumerate}

\section{Related work}
\label{sec:related}

\paragraph{Sphere uniformity.} \citet{wang2020understanding} proposed
a pairwise repulsive log-sum-exp kernel on $\Sd$ matching the
contrastive fixed point. The construction is geometric rather than
probabilistic: it does not target $\mathcal{N}(0, I_d)$, and
concatenated uniform-on-sphere blocks are not uniform on the larger
sphere because the per-block energies are pinned at $\|w_1\|^2 =
\|w_2\|^2 = 1/2$, while the truly uniform split follows a Beta
distribution, breaking compositionality of independent factor encoders.

\paragraph{Variance-covariance regularization.}
VICReg~\citep{bardes2022vicreg} regularizes variance and covariance
--- first- and second-order statistics only.
\citet{kuang2025radial} add a Vasicek $m$-spacing KL between the
empirical norm and Chi$(d)$. Their Proposition~1 shows the radial
map Gaussianizes exactly the elliptically symmetric distributions
--- a strict superset of Gaussians, but \emph{not all distributions}:
the X-shaped 2D distribution with identity covariance is a
counterexample. The wristband loss is biconditional with
$\mathcal{N}(0, I_d)$ on \emph{arbitrary} distributions on $\R^d
\setminus \{0\}$.

\paragraph{Density-matching at the batch level.} Maximum Mean
Discrepancy~\citep{gretton2012kernel} (and the Wasserstein
autoencoder~\citep{tolstikhin2018wasserstein} built on it) is
theoretically sound but bandwidth- and dimension-sensitive in
practice; sliced Wasserstein~\citep{kolouri2019sliced} is
characteristic at the population level but, with finite projections,
can miss localized joint clumping and ignore the radial geometry of
Gaussians. Hungarian-style optimal transport gives strong evaluations
but is too expensive for the training loop.

\paragraph{Stochastic alternatives.} VAEs~\citep{kingma2014auto} and $\beta$-VAEs~\citep{higgins2017beta} replace per-sample distribution-matching with a per-sample KL, requiring sampling on every forward pass and suffering from posterior collapse. Diffusion~\citep{ho2020denoising} and flow-matching~\citep{lipman2023flow} learn a transport from noise to data via a multi-step ODE/SDE; they produce excellent samples but cannot return a deterministic latent code in a single forward pass. In contrast, the wristband loss evaluates a static batch energy in a single deterministic forward pass.

\paragraph{Normalizing flows.} RealNVP~\citep{dinh2017density} introduced exactly invertible affine coupling layers; we use them as a postprocessing stage between the encoder and the wristband loss to give the encoder freedom to choose any intermediate parameterization while the flow handles the final shaping toward $\mathcal{N}(0, I)$.

\section{The wristband map}
\label{sec:wristband}

\subsection{Definition and equivalence theorem}

Let $d \ge 2$ and let $Q$ be a Borel probability measure on $\R^d$ with $Q(\{0\})=0$. Define the \emph{wristband map}
\begin{equation}
  W = \Sd \times [0, 1]
  \qquad
  \Phimap : \R^d \setminus \{0\} \to W, 
  \qquad
  \Phimap(x) \;=\; \left(\, \frac{x}{\|x\|},\; F_{\chi^2_d}\!\left(\|x\|^2\right) \,\right),
  \label{eq:wristband-map}
\end{equation}
where $F_{\chi^2_d}(s) = \gamma(d/2, s/2) / \Gamma(d/2)$ is the CDF of the chi-squared distribution with $d$ degrees of freedom. Write $\sigma_{d-1}$ for the uniform probability measure on $\Sd$ and $\unif[0,1]$ for Lebesgue measure on $[0, 1]$.

\begin{theorem}[Wristband equivalence]
  \label{thm:equivalence}
  For $d \ge 2$ and $Q$ as above,
  \begin{equation}
    \Phimap_\# Q \;=\; \sigma_{d-1} \otimes \unif[0, 1]
    \quad \Longleftrightarrow \quad
    Q \;=\; \mathcal{N}(0, I_d).
  \end{equation}
\end{theorem}

Full proof in App.~\ref{app:proof}; both directions are machine-verified in Lean~4 (\S\ref{sec:formalization}). The analogous $d=1$ statement holds with the Rademacher sign on $S^0=\{-1,+1\}$; we state $d\ge2$ because the continuous angular machinery requires it.

\subsection{Composability}
\label{sec:composability}

A direct corollary makes the framework attractive for modular models. Suppose $E_a: \R^{d_a^{\text{in}}} \to \R^{d_a}$ and $E_b: \R^{d_b^{\text{in}}} \to \R^{d_b}$ are encoders trained so that $E_a(X_a) \sim \mathcal{N}(0, I_{d_a})$ and $E_b(X_b) \sim \mathcal{N}(0, I_{d_b})$, and that the \emph{joint} concatenated latent satisfies
\[
(E_a(X_a), E_b(X_b)) \sim \mathcal{N}(0, I_{d_a+d_b}).
\]
Then Gaussian factorization gives $E_a(X_a) \indep E_b(X_b)$. At inference time the encoder $E_b$ may therefore be replaced by a draw $\varepsilon \sim \mathcal{N}(0, I_{d_b})$ to estimate any functional of a downstream predictor under resampling of that factor. The joint condition is essential: marginal Gaussianity of each block is not enough, since two standard-normal blocks can remain statistically dependent.

This exact independence statement should not be read as a promise that arbitrary dependent observations can be deterministically encoded into independent lossless blocks. If $X_a$ and $X_b$ share mutual information and both encoders must preserve that shared information for reconstruction, an exactly independent deterministic split is impossible without discarding information. In dependent-factor settings we instead use an asymmetric context/residual construction: the observed factor is encoded into a deterministic \emph{context} block carrying the information inferable from evidence, while a smaller Gaussian \emph{residual} block carries the remaining uncertainty. Section~\ref{sec:exp-mnist} demonstrates this construction on MNIST top-half sampling conditioned on the bottom half.

\section{The wristband Gaussian loss}
\label{sec:loss}

We now derive a finite-sample loss whose unique population minimum is $\mathcal{N}(0, I_d)$; the core term is the reflected pairwise repulsion in \S\ref{sec:reflection}, while the remaining components are optional accelerators that share the same optimum. Let $\{x_i\}_{i=1}^N \subset \R^d$ be a batch of point embeddings, and write $u_i = x_i / \|x_i\|$, $s_i = \|x_i\|^2$, $t_i = F_{\chi^2_d}(s_i)$.

\subsection{The wristband repulsion kernel}
\label{sec:reflection}
The main term is a soft pairwise repulsion in $(u,t)$-space. 
\begin{equation}
  K_W(u, t;\, u', t') \;=\; k_{\mathrm{ang}}(u, u') \cdot k_{\mathrm{rad}}(t,t')
  \label{eq:kernel}
\end{equation}

The angular component of the kernel is the chordal Gaussian:
\begin{equation}
  k_{\mathrm{ang}}(u,u')
  = \exp\!\left\{-\beta\alpha^2\|u-u'\|^2\right\}
  = e^{-2\beta\alpha^2}\exp\!\left\{(2\beta\alpha^2)u^\top u'\right\},
  \label{eq:ang-kernel}
\end{equation}
here $\alpha$ balances the radial and angular scales. $\beta > 0$ sets the radial interaction range. Small $\beta$ gives a soft repulsion; large $\beta$ gives a sharp, short-range one.

Let $t,t'\in[0,1]$. The infinite-image Neumann reflection of the radial Gaussian is
\begin{equation}
k_{\mathrm{rad}}(t,t')
=
\sum_{m\in\mathbb{Z}} e^{-\beta(t-t'-2m)^2}
+
\sum_{m\in\mathbb{Z}} e^{-\beta(t+t'-2m)^2}.
\label{eq:rad-infinite}
\end{equation}

Define the three-image reflected kernel, truncated from the expression above:
\begin{equation}
  k_{\mathrm{img}}(t,t') \;=\;
  e^{-\beta(t-t')^2}
  + e^{-\beta(t+t')^2}
  + e^{-\beta(t+t'-2)^2}.
  \label{eq:rad-3image}
\end{equation}
The three radial terms correspond to the real point $t'$, its reflection $-t'$ across the boundary $0$, and its reflection $2-t'$ across the boundary $1$.

The implemented pairwise repulsion uses $K^{\mathrm{img}}_W(u,t;\,u',t')=k_{\mathrm{ang}}(u,u')\,k_{\mathrm{img}}(t,t')$:
\begin{equation}
  \Lrep
  =
  \frac{1}{\beta}
  \log\!\left(
  \frac{\sum_{i,j=1}^N K^{\mathrm{img}}_W(u_i,t_i;\,u_j,t_j)-N}{3N^2-N}
  + \epsilon
  \right).
  \label{eq:repulsion}
\end{equation}
The subtraction removes the $N$ real self-interactions $(i=j,t_i=t_j)$ whose kernel value is exactly one. The two reflected self-images are intentionally retained: they are the finite-batch boundary-correction terms that prevent the radial coordinate from exploiting the missing mass outside $[0,1]$.

\paragraph{Why reflection.} Reflection restores the missing kernel mass near the boundaries of $[0,1]$, yielding the zero-flux behavior appropriate for a uniform target. The three-image form in \eqref{eq:rad-3image} is the nearest-image truncation of the Neumann sum \eqref{eq:rad-infinite}; the next omitted image contributes at most $e^{-\beta}$ per term (e.g., $3.4\times10^{-4}$ at $\beta=8$).

For a Borel probability measure $P$ on $W$, define $\mathcal{E}(P)=\iint_{W\times W}K_W(w,w')\,dP(w)\,dP(w')$.

\begin{theorem}[Energy minimization]
  \label{thm:minimization}
  For the kernel \eqref{eq:kernel} with $k_{\mathrm{rad}}$ as in \eqref{eq:rad-infinite}, $\beta,\alpha > 0$, and $d \ge 2$, every Borel probability measure $P$ on $W$ satisfies
  \begin{equation}
    \mathcal{E}(P) \;\ge\; \mathcal{E}(\sigma_{d-1} \otimes \unif[0, 1]),
  \end{equation}
  with equality iff $P = \sigma_{d-1} \otimes \unif[0, 1]$.
\end{theorem}
Combined with Theorem~\ref{thm:equivalence}, this gives a population-level guarantee: minimizing the wristband repulsion drives $\Phimap_\# Q \to \sigma_{d-1}\otimes\unif[0,1]$, and hence $Q \to \mathcal{N}(0,I_d)$. Theorem~\ref{thm:minimization} is also machine-verified in Lean~4 (\S\ref{sec:formalization}).

\subsection{Optional accelerators: radial Wasserstein and moment terms}

The full infinite-image reflected repulsion is the only term required for the population characterization (Theorems~\ref{thm:equivalence}--\ref{thm:minimization}; cf.~\S\ref{sec:formalization}). In finite-sample training it can be slow to correct global drift, so we optionally add two auxiliary terms that share the same target but accelerate convergence:
\begin{align}
  \Lrad &= \frac{1}{N} \sum_{i=1}^N \left( t_{(i)} - \tfrac{i - 1/2}{N} \right)^2,
  \label{eq:radial}\\
  \Lmom &= W_2^2\!\left( \mathcal{N}(\hat{\mu}, \hat{\Sigma}),\, \mathcal{N}(0, I_d) \right) \;=\; \|\hat{\mu}\|^2 + \sum_{i=1}^d \left( \sqrt{\hat{\lambda}_i} - 1 \right)^2,
  \label{eq:moment}
\end{align}
where $t_{(i)}$ are the order statistics of $\{t_i\}$ and $\hat{\lambda}_i$ are the eigenvalues of $\hat{\Sigma}$. $\Lrad$ is a 1D squared-Wasserstein distance to $\unif[0, 1]$, and $\Lmom$ is the closed-form 2-Wasserstein distance between the batch's fitted Gaussian and the target. Both vanish at the wristband target.

\subsection{Self-calibration via \texorpdfstring{$z$}{z}-scoring}
\label{sec:calibration}

Each individual term has a scale that depends on $N$, $d$, and the bandwidth $\beta$. Rather than tune relative weights per problem, we draw $M$ batches from $\mathcal{N}(0, I_d)$ at construction time --- once --- and record the empirical mean $\bar{\mu}_*$ and standard deviation $\bar{s}_*$ of each component $* \in \{\mathrm{rep}, \mathrm{rad}, \mathrm{mom}\}$. Define
\[
  \widetilde{\mathcal{L}}_*
  =
  \frac{\mathcal{L}_*(\{x_i\})-\bar{\mu}_*}{\bar{s}_*}.
\]
The training-time numerator is
\[
  S
  =
  w_{\mathrm{rep}}\widetilde{\Lrep}
  + w_{\mathrm{rad}}\widetilde{\Lrad}
  + w_{\mathrm{mom}}\widetilde{\Lmom}.
\]
Finally, we divide by the Monte-Carlo standard deviation $\bar{s}_{S}$ of this same numerator under the null:
\begin{equation}
  \Lwb \;=\; \frac{S}{\bar{s}_{S}}.
  \label{eq:zscore}
\end{equation}
This directly accounts for correlations among the calibrated components; it does not assume that they are independent. Under the null hypothesis $\{x_i\} \overset{\text{i.i.d.}}{\sim} \mathcal{N}(0,I_d)$, $\Lwb$ has mean approximately $0$ and standard deviation approximately $1$ by construction.

\section{Spectral Neumann}
\label{sec:spectral}
The pairwise loss \eqref{eq:repulsion} costs $O(N^2 d)$ per step. For large batches we provide a spectral fast path in $O(NdK)$ time and $O(dK)$ memory: the three-image radial kernel is replaced by its infinite-image Neumann form, truncated to $K$ cosine modes and angular degrees $\ell \in \{0,1\}$. 

\subsection{Angular eigenvalues}
From \eqref{eq:ang-kernel} and as a continuous zonal kernel on $\Sd$ under the probability measure $\sigma_{d-1}$, our angular kernel admits the Mercer expansion in the spherical-harmonic basis as in \citet{dutordoir2020sparse},
\begin{equation}
k_{\mathrm{ang}}(u,u')
=
\sum_{\ell=0}^{\infty}
\lambda_\ell
\sum_{m=1}^{N_\ell}Y_{\ell,m}(u)Y_{\ell,m}(u'),
\label{eq:ang-mercer}
\end{equation}
where $\{Y_{\ell,m}\}_{m=1}^{N_\ell}$ is an orthonormal basis for the degree-$\ell$ spherical harmonics and $N_\ell$ is its multiplicity. The Funk-Hecke theorem gives $\lambda_\ell$ as the Gegenbauer coefficient of the kernel function \citep{dai2013analysis}, by Gegenbauer's extension of Poisson's integral for the Bessel function \citep{demicheli2018integral}:
\begin{equation}
  \lambda_\ell
  =
  \Gamma(\nu + 1)
  \left(\frac{2}{c}\right)^{\nu}
  e^{-c} I_{\nu+\ell}(c),
  \qquad
  \nu=\frac{d-2}{2},
  \qquad
  c=2\beta\alpha^2,
  \label{eq:angular-eigs}
\end{equation}
where $I_\rho$ is the modified Bessel function of the first kind. Multiplicity is handled by the sum over $m$ in \eqref{eq:ang-mercer}. In particular, the degree-one eigenspace has the explicit orthonormal basis
\[
Y_{1,p}(u)=\sqrt{d}\,u_p,
\qquad p=1,\ldots,d,
\]
since $\E_{\sigma_{d-1}}[u_pu_q]=\delta_{pq}/d$.

\subsection{Radial Neumann cosine modes}

By Poisson summation, using the unnormalized cosine convention of our implementation,
\begin{equation}
k_{\mathrm{rad}}(t,t')
=
\sum_{k=0}^{\infty}
a_k\cos(k\pi t)\cos(k\pi t'),
\label{eq:rad-cos}
\end{equation}
with
\begin{equation}
a_0=\sqrt{\frac{\pi}{\beta}},
\qquad
a_k=2\sqrt{\frac{\pi}{\beta}}\exp\!\left\{-\frac{\pi^2k^2}{4\beta}\right\}
\quad (k\ge1).
\label{eq:rad-coeffs}
\end{equation}
The factor of $2$ for $k\ge1$ is exactly the conversion from the orthonormal basis $\sqrt{2}\cos(k\pi t)$ to the unnormalized basis $\cos(k\pi t)$.

\subsection{The truncated spectral energy}
Multiplying \eqref{eq:ang-mercer} and \eqref{eq:rad-cos} and re-indexing as a single double sum over $(\ell,m,k)$ gives the infinite-image kernel as a tensor product of $\{Y_{\ell,m}(u)\cos(k\pi t)\}$ modes with eigenvalues $\lambda_\ell a_k$. Truncating the angular expansion to degrees $\ell\in\{0,1\}$ and the radial expansion to $K$ cosine modes ($k=0,\dots,K-1$), we define the following summary statistics for a batch of $N$ points:
\begin{equation}
c_{0,k}=\frac{1}{N}\sum_{i=1}^{N}\cos(k\pi t_i),
\qquad
c_{1,k}=\frac{\sqrt{d}}{N}\sum_{i=1}^{N}u_i\cos(k\pi t_i)\in\R^d.
\end{equation}
The truncated Spectral Neumann kernel energy is:
\begin{equation}
\mathcal{E}_{sp}
=
\lambda_0\sum_{k=0}^{K-1}a_k c_{0,k}^2
+
\lambda_1\sum_{k=0}^{K-1}a_k\|c_{1,k}\|^2,
\label{eq:spectral-energy}
\end{equation}
where the respective spectral repulsion becomes
\begin{equation}
\Lrep^{\mathrm{sp}}
=
\frac{1}{\beta}
\log\!\left(\frac{\mathcal{E}_{sp}}{\lambda_0a_0}+\epsilon\right),
\label{eq:spectral-rep}
\end{equation}
where $\lambda_0a_0$ is the product-uniform population value of the retained constant mode. The compute pattern is a single forward pass over the batch with $O(NdK)$ work and $O(dK)$ state; pseudocode is given in App.~\ref{app:spectral-pseudo}. The timing study in \S\ref{sec:exp-spectral} also reports a leaner $K=3$ setting to expose the speed/fidelity trade-off.

\section{Lean 4 formalization}
\label{sec:formalization}

Theorems~\ref{thm:equivalence} and~\ref{thm:minimization} are machine-checked in Lean~4 (\texttt{mathlib} v4.28.0). The wristband equivalence (\texttt{wristbandEquivalence}) is \texttt{sorry}-free as an iff; energy minimization and uniqueness are established for the infinite-image Neumann kernel modulo a small inventory of imported textbook axioms and routine analytic support lemmas.

\section{The Deterministic Gaussian Autoencoder (DGAE)}
\label{sec:dgae}

The wristband loss can be combined with any encoder. We pair it with two building blocks that empirically help training. \textbf{Learnable-key Euclidean attention} is a multi-head softmax layer with free-parameter keys $\{k_j\}$ and values $\{v_j\}$ (not projections of input tokens); the Euclidean logits with inverse-temperature $\tau$ are $\ell_j(q)=\tau(q^\top k_j-\tfrac{1}{2}\|k_j\|^2) = -\tfrac{\tau}{2}\|q-k_j\|^2$ up to a query-dependent constant that cancels inside the softmax, so attention becomes soft nearest-RBF-prototype retrieval. \textbf{An exact invertible flow} $f$ (a stack of RealNVP coupling layers with deterministic permutations and identity initialization) sits between the encoder and the loss: the forward map shapes the code into the wristband target, the decoder uses the exact inverse, and no reconstruction signal is wasted on regularization pressure. The total training loss for data $\{x_i\}$, encoder $E$, decoder $D$ is
\begin{equation}
  \mathcal{L}(\{x_i\}) \;=\; \lambda_{\mathrm{rec}} \cdot \frac{1}{N}\sum_{i=1}^N \|D(f^{-1}(z_i)) - x_i\|^2 \;+\; \lambda_{\mathrm{wb}} \cdot \Lwb(\{z_i\}),
  \quad z_i = f(E(x_i)).
\end{equation}
There is no sampling, no KL term, no reparameterization trick, no $\beta$-anneal. The same input always maps to the same latent code and the same reconstruction.

\section{Experiments}
\label{sec:experiments}

We evaluate four properties: (i)~Gaussianity recovery on two synthetic direct point-cloud benchmarks without any encoder or flow; (ii)~end-to-end training of a DGAE on a strongly non-Gaussian 15-D mixture; (iii)~conditional sampling on MNIST as an end-to-end test of the dependent-factor context/residual construction; and (iv)~spectral vs.\ pairwise wall-clock and gradient-parity trade-offs.

\subsection{Gaussianity recovery on the X-distribution}
\label{sec:exp-x}

The X-distribution of \citet{kuang2025radial} is a 2D mixture concentrated on two crossing line segments. We use the dimension-$d$ axis-uniform generalization: choose a coordinate axis uniformly, choose a signed coordinate uniformly in $[-\sqrt{3d},\sqrt{3d}]$, and set all other coordinates to zero. This distribution has identity covariance but is not elliptically symmetric; in 2D it is the same crossing-line construction up to rotation. It separates the objectives cleanly: VCReg sees nearly correct moments, Radial-VCReg can repair the radius distribution, and Wristband targets the full joint direction--radius law that is equivalent to Gaussianity.

We optimize $N = 4096$ point variables directly, with no encoder, decoder, or flow. VCReg uses SGD with initial learning rate $0.02$; Radial-VCReg uses the official synthetic radial weights $100/100$ with Adam learning rate $0.1$; MMD uses multiscale Gaussian bandwidths $\sigma=\sqrt{d}\{0.25,0.5,1,2,4\}$; and sliced $W_2$ uses $128$ random projections. Wristband uses Adam learning rate $0.05$, global reduction, $\lambda_{\mathrm{rad}}=0.1$, $\lambda_{\mathrm{mom}}=1$, $\beta=64$, and $\alpha=0.8$. All methods run for $5000$ optimization steps.

The evaluation metric is calibrated barycentric $W_2$. For each $(N,d)$, we generate $128$ Gaussian batches, randomly pair them, solve exact Hungarian matchings within each pair, replace each pair by the midpoint batch of matched points, and recurse until one deterministic Gaussian reference batch remains. We then compute exact equal-weight $W_2$ distance from each optimized batch to this reference and report the resulting $z$-score against $128$ fresh Gaussian batches of the same shape. Results are reported in Table~\ref{tab:x-distribution}; lower is better, $z=0$ is the finite-sample Gaussian null, and $z<0$ indicates a lower-discrepancy point set than a typical i.i.d.\ Gaussian batch of the same shape. See Appendix~\ref{app:sunshine-dist} for a visual example comparing Radial-VCReg to Wristband.

\begin{table}[t]
  \caption{Barycentric $W_2$ $z$-score on axis-uniform X distributions after $5000$ direct-sample optimization steps. Lower is better; $0$ is the Gaussian finite-sample null. Mean $\pm$ s.d.\ over 5 seeds.}
  \label{tab:x-distribution}
  \centering
  \begin{tabular}{lccc}
    \toprule
    Loss & 2D $z_{W_2}$ & 10D $z_{W_2}$ & optimized signal \\
    \midrule
    None (raw X-distribution) & $120.15 \pm 0.78$ & $138.99 \pm 2.09$ & none \\
    VCReg \citep{bardes2022vicreg} & $120.24 \pm 0.65$ & $139.15 \pm 2.06$ & moments \\
    Radial-VCReg \citep{kuang2025radial} & $48.41 \pm 1.93$ & $76.83 \pm 0.97$ & moments + radius \\
    MMD \citep{gretton2012kernel} & $4.30 \pm 1.00$ & $-5.77 \pm 0.63$ & kernel match \\
    Sliced $W_2$ \citep{kolouri2019sliced} & $\mathbf{-3.80 \pm 0.49}$ & $-3.46 \pm 1.37$ & projections \\
    \textbf{Wristband (ours)} & $-1.12 \pm 0.82$ & $\mathbf{-10.52 \pm 0.38}$ & wristband \\
    \bottomrule
  \end{tabular}
\end{table}

\subsection{Radial--angular copula impostor}
\label{sec:exp-rac}

A spherical Gaussian has the polar decomposition $X=RU$, where $R\sim\chi_d$, $U\sim\sigma_{d-1}$, and $R\indep U$. The X benchmark violates the radial marginal, which makes it useful but not maximally adversarial. We therefore also test a radial--angular copula (RAC) impostor that has the correct Gaussian radius law and the correct uniform direction marginal by construction, but breaks the independence between them.

For each batch, we sample directions $U_i$ uniformly on $\Sd$ and radii $R_i$ independently from $\chi_d$. We compute the angular entropy score
\[
  s(U_i)=-\sum_{j=1}^d U_{ij}^2\log(U_{ij}^2+\epsilon),
\]
sort the radii and the scores, and assign the smallest radius to the smallest score, continuing rankwise. This preserves both empirical marginals exactly in the finite batch while inducing a comonotone copula between radius and direction. We use $N=2048$, dimensions $d\in\{10,128\}$, the same fixed-step protocol and evaluation metric as above, and the same Wristband, MMD, and sliced-$W_2$ settings. For Radial-VCReg we use the stable parser-default learning rate $5\times 10^{-3}$ for this fixed-step protocol.

\begin{table}[t]
  \caption{Barycentric $W_2$ $z$-score on the RAC Gaussian impostor. The data have the correct Gaussian radial and angular marginals but the wrong radial--angular copula. Lower is better; $0$ is the Gaussian finite-sample null. Mean $\pm$ s.d.\ over 5 seeds.}
  \label{tab:rac}
  \centering
  \begin{tabular}{lcc}
    \toprule
    Loss & 10D $z_{W_2}$ & 128D $z_{W_2}$ \\
    \midrule
    VCReg \citep{bardes2022vicreg} & $0.82 \pm 0.35$ & $4.36 \pm 0.27$ \\
    Radial-VCReg \citep{kuang2025radial} & $0.68 \pm 0.25$ & $0.22 \pm 0.12$ \\
    MMD \citep{gretton2012kernel} & $-3.42 \pm 1.04$ & $-0.99 \pm 0.63$ \\
    Sliced $W_2$ \citep{kolouri2019sliced} & $-1.96 \pm 1.09$ & $-0.28 \pm 1.18$ \\
    \textbf{Wristband (ours)} & $\mathbf{-7.44 \pm 0.11}$ & $\mathbf{-1.18 \pm 0.07}$ \\
    \bottomrule
  \end{tabular}
\end{table}

\subsection{Deterministic Gaussian Autoencoder on a 15-D non-Gaussian mixture}
\label{sec:exp-gae}

We follow the synthetic benchmark in the open-source companion code: $100{,}000$ samples in $\R^{15}$ drawn from a 5-cluster mixture of anisotropic Gaussians, then standardized. The encoder is a 64-head, 256-basis Euclidean attention module with an Auto-Compressing Network~\citep{dorovatas2025auto} head combiner; the decoder is the same learnable-key attention module with input/output dimensions reversed; the flow has 4 affine coupling layers; embedding dimension $d = 8$; batch size $1024$; 40 epochs; $\lambda_{\mathrm{rec}} = 1$, $\lambda_{\mathrm{wb}} = 0.1$. Full architectural details are in Appendix~\ref{app:hyper}.

Final reconstruction MSE is $0.241 \pm 0.006$. The wristband total $\Lwb$ on the trained latents drops from $3405.1 \pm 88.2$ at initialization to $6.28 \pm 1.79$ after training, showing a large reduction. The final latent coordinate scales are near one.

\subsection{Conditional sampling on MNIST: dependent factors}
\label{sec:exp-mnist}

Each MNIST digit is split into top and bottom $14\times28$ halves $(x_{\mathrm{top}},x_{\mathrm{bot}})$ -- a dependent-factor setting where the bottom strongly constrains the plausible top. We instantiate the context/residual construction from \S\ref{sec:composability}: the bottom encoder produces a context latent $z_{\mathrm{bot}}\in\R^{18}$, the top encoder a residual latent $z_{\mathrm{top}}\in\R^{3}$, each with its own invertible flow, and the wristband loss is applied to the joint $z=[z_{\mathrm{bot}},z_{\mathrm{top}}]\in\R^{21}$. Training combines a full mode ($\hat{x}_{\mathrm{full}}=D(z_{\mathrm{bot}},z_{\mathrm{top}})$) with a conditional mode that encodes only the bottom half and averages a pixelwise $L_1$ reconstruction loss over $K{=}32$ residual draws (\eqref{eq:mnist-cf} in App.~\ref{app:hyper}); $\mathcal{L}_{\mathrm{cf}}$ acts as a weak regularizer ($\lambda_{\mathrm{cf}}=0.1$) after a full-reconstruction warm-up, with weights $\lambda_{\mathrm{rec}}=1$, $\lambda_{\mathrm{wb}}=0.1$.

Across five seeds, the measured test full-reconstruction loss is $0.0369 \pm 0.0007$ in pixelwise $L_1$, the conditional expected loss is $0.0533 \pm 0.0008$, and the wristband total is $1.90 \pm 0.33$. At inference, only $x_{\mathrm{bot}}$ is encoded; $z_{\mathrm{top}}$ is replaced by fresh draws from $\mathcal{N}(0,I_3)$. This produces multiple completions for the same observed bottom half. For qualitative inpainting grids, the observed bottom half is clamped back into the displayed image after decoding so that the visualization isolates variation in the sampled top half. This experiment validates the deterministic context/residual interface; calibrated posterior generation on stronger image benchmarks is orthogonal to the Gaussianization loss itself.

\begin{figure*}[t]
  \centering
  \includegraphics[width=\textwidth]{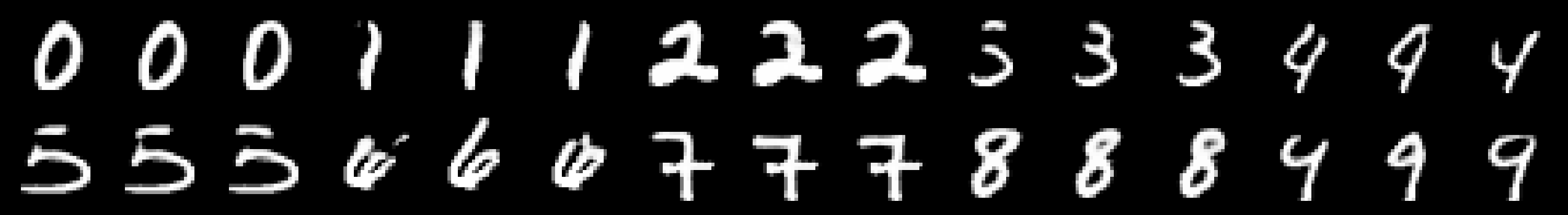}
  \caption{MNIST conditional sampling grid. The bottom half is observed and clamped for display; the top half is decoded from fresh residual samples. Note how the bottom of 4 and 9 is not distinctive and the sampled top reflects this. See Appendix~\ref{app:mnist-grid-full} for more examples.}
  \label{fig:mnist-grid}
\end{figure*}

\subsection{Spectral vs.\ pairwise: speedup and parity}
\label{sec:exp-spectral}

Table~\ref{tab:spectral-parity} reports the spectral-vs-pairwise
speed/fidelity trade-off across embedding dimension and batch size.
We evaluate on four structured non-Gaussian generators Appendix~\ref{app:hyper}:
a standardized Gaussian mixture, an axis-aligned two-mode mixture,
an i.i.d.\ heavy-tailed Student-$t$, and a ring-shaped distribution
with clustered angles. We use a lean $K=3$ radial truncation here so
the approximation stays distinguishable from the pairwise path.
Value correlations exceed $0.99998$ in all rows; we therefore report
input-gradient cosine similarity and compiled forward+backward
timing. A DGAE run with default spectral loss gives reconstruction
MSE $0.231\pm0.026$ vs $0.241\pm0.006$ (pairwise), with mean
absolute seedwise MSE difference $0.022\pm0.013$.

\begin{table}[t]
  \caption{Spectral Neumann ($K = 3$ radial modes) vs.\ exact pairwise repulsion on four structured non-Gaussian batches and 5 seeds. Timing is compiled forward+backward wall-clock on an NVIDIA RTX PRO 6000 Blackwell Workstation Edition. Higher is better for gradient cosine and speedup.}
  \label{tab:spectral-parity}
  \centering
  \begin{tabular}{rrcccc}
    \toprule
    $d$ & $N$ & mean grad cos & min grad cos & pair/spec ms & speedup \\
    \midrule
    16  & $1024$ & $0.9821$ & $0.9152$ & $0.63/0.57$ & $1.10\times$ \\
    16  & $8192$ & $0.9882$ & $0.9378$ & $2.96/0.56$ & $5.32\times$ \\
    64  & $1024$ & $0.9732$ & $0.9123$ & $0.63/0.58$ & $1.08\times$ \\
    64  & $8192$ & $0.9844$ & $0.9075$ & $3.17/0.56$ & $5.69\times$ \\
    256 & $1024$ & $0.9958$ & $0.9822$ & $0.65/0.59$ & $1.09\times$ \\
    256 & $8192$ & $0.9796$ & $0.9095$ & $3.46/0.55$ & $6.26\times$ \\
    \bottomrule
  \end{tabular}
\end{table}

\section{Discussion and limitations}

\paragraph{Discussion.} The consequential difference from existing radial-Gaussianization methods is the move from a \emph{necessary} condition (matching the radial marginal) to a \emph{necessary and sufficient} one (joint uniformity on $\Sd \times [0, 1]$): the population target is uniquely $\mathcal{N}(0, I_d)$ on arbitrary distributions, not only elliptically symmetric ones. The X-distribution of \S\ref{sec:exp-x} separates moment and radial objectives from full Gaussianization, and the RAC benchmark of \S\ref{sec:exp-rac} isolates the harder case where both the radial and angular marginals are already correct but their copula is wrong. This biconditional is enforced by $\Lrep$ alone; the radial and moment penalties are included only to improve finite-sample optimization, and the invertible flow is an architectural aid that lets the encoder choose a task-useful intermediate parameterization while the flow handles the final shaping toward the Gaussian interface. The direct point-cloud experiments isolate the loss itself with no encoder, decoder, or flow. In these direct benchmarks we use the default $\beta=64$, $\alpha=0.8$, and global reduction; neural experiments use the smaller-kernel settings stated in Appendix~\ref{app:hyper}. Sensitivity sweeps in $\beta\in[4,16]$ change the 15D DGAE reconstruction MSE by at most $8.9\%$.

\paragraph{Limitations.} First, both kernel paths approximate the
infinite-image reflected kernel: the three-image path truncates the
image sum (next-image error $O(e^{-\beta})$); the spectral path
truncates radial/angular modes and uses a V-statistic, with gradient
parity validated empirically (\S\ref{sec:exp-spectral}). Second, it
requires $d\ge3$ and the chordal angular kernel; geodesic-angular
and 1D embeddings remain on the pairwise path. Third, the wristband
map is undefined at $x=0$; the implementation's numerical floor is
a formal discontinuity, benign in our experiments. Fourth, exact
block independence requires independent underlying information; the
MNIST context/residual construction handles dependent factors
without exact posterior recovery.

\section{Conclusion}

We presented the wristband Gaussian loss, a deterministic, single-pass, GPU-friendly batch loss whose population target is uniquely $\mathcal{N}(0,I_d)$ for arbitrary distributions on $\R^d\setminus\{0\}$ with $d\ge2$. The construction rests on a necessary-and-sufficient sphere--interval decomposition of the Gaussian, a principled reflected-kernel correction for the bounded radial coordinate, and a null calibration that makes component scales comparable. The pairwise path is the reference loss; an optional spectral Neumann fast path reduces the large-batch cost from $O(N^2d)$ to $O(NdK)$ as an explicit approximation. Coupled with an invertible flow and learnable-key Euclidean attention, the resulting deterministic Gaussian autoencoder produces plug-compatible Gaussian latents that admit counterfactual sampling at inference.

\appendix

\section{Proof of Theorem~\ref{thm:equivalence}}
\label{app:proof}

We provide the standard measure-theoretic proof; the corresponding statement is formalized in Lean~4 in the accompanying development.

\paragraph{($\Leftarrow$) $Q = \mathcal{N}(0, I_d) \implies \Phimap_\# Q = \sigma_{d-1} \otimes \unif[0, 1]$.}
Let $X\sim\mathcal{N}(0,I_d)$. Write $R=\|X\|$ and $U=X/\|X\|$. The Gaussian polar decomposition gives $U\sim\sigma_{d-1}$, $R^2\sim\chi^2_d$, and $U\indep R$. By the probability integral transform, $T=F_{\chi^2_d}(R^2)\sim\unif[0,1]$, and since $T$ is a measurable function of $R^2$, we also have $U\indep T$. Therefore $\Phimap(X)=(U,T)\sim\sigma_{d-1}\otimes\unif[0,1]$.

\paragraph{($\Rightarrow$) $\Phimap_\# Q = \sigma_{d-1} \otimes \unif[0, 1] \implies Q = \mathcal{N}(0, I_d)$.}
Let $X\sim Q$ and $(U,T)=\Phimap(X)$. By assumption, $U\sim\sigma_{d-1}$, $T\sim\unif[0,1]$, and $U\indep T$. Define $S=F_{\chi^2_d}^{-1}(T)$ and $R=\sqrt{S}$. By the reverse probability integral transform, $S\sim\chi^2_d$, so $R$ has the Chi$(d)$ density
\[
f_R(r)
=
\frac{1}{2^{d/2-1}\Gamma(d/2)}
r^{d-1}e^{-r^2/2},
\qquad r>0.
\]
Moreover, $R$ is a measurable function of $T$, hence $R\indep U$. Since $X=RU$ almost surely by construction of $\Phimap$, $(R,U)$ are the polar coordinates of $X$. Let $\omega_{d-1}=2\pi^{d/2}/\Gamma(d/2)$ be the surface area of $\Sd$. The joint density of $(R,U)$ is $f_R(r)/\omega_{d-1}$, and the change of variables $x=ru$ has Jacobian $r^{d-1}$. Thus the induced density of $X$ is
\[
f_X(x)
=
\frac{f_R(\|x\|)}
{\omega_{d-1}\|x\|^{d-1}}
=
\frac{1}{(2\pi)^{d/2}}
e^{-\|x\|^2/2},
\]
which is exactly the $\mathcal{N}(0,I_d)$ density.

\paragraph{Remark on $d=1$.}
The same probabilistic equivalence holds in one dimension if the direction is interpreted as the sign $U=\mathrm{sign}(X)\in S^0=\{-1,+1\}$. We state the main theorem for $d\ge2$ because the continuous angular repulsion and spherical-harmonic spectral approximation are the objects used by the loss.

\section{Spectral Neumann pseudocode}
\label{app:spectral-pseudo}

\begin{algorithm}[H]
  \caption{Spectral Neumann wristband repulsion (single forward pass; global reduction)}
  \label{alg:spectral}
  \begin{algorithmic}[1]
    \Require batch $\{x_i\}_{i=1}^N \subset \R^d$; precomputed $(\lambda_0,\lambda_1,a_0,\ldots,a_{K-1})$.
    \State $u_i \gets x_i/\|x_i\|$, $\;t_i \gets F_{\chi^2_d}(\|x_i\|^2)$ \Comment{$O(Nd)$}
    \State $C_{i,k}\gets \cos(k\pi t_i)$ for $k=0,\ldots,K-1$ \Comment{$O(NK)$}
    \State $c_{0,k}\gets N^{-1}\sum_i C_{i,k}$ \Comment{$O(NK)$}
    \State $c_{1,k}\gets (\sqrt{d}/N)\sum_i u_i C_{i,k}$ \Comment{$O(NdK)$}
    \State $\mathcal{E}_K\gets \lambda_0\sum_k a_k c_{0,k}^2+\lambda_1\sum_k a_k\|c_{1,k}\|^2$
    \State \Return $\Lrep^{\mathrm{spec}}\gets \beta^{-1}\log(\mathcal{E}_K/(\lambda_0a_0)+\epsilon)$
  \end{algorithmic}
\end{algorithm}

\section{Calibration constants}

Table~\ref{tab:calibration} reports the Monte-Carlo null statistics $(\bar{\mu}_*, \bar{s}_*)$ used in $z$-scoring, at the canonical batch size and dimension. We use \texttt{calibration\_reps}=4096 batches drawn from $\mathcal{N}(0, I_d)$. On the NVIDIA RTX PRO 6000 Blackwell Workstation Edition used for the experiments, the measured calibration-only times were 4.93s for $d=8$, 8.76s for $d=64$, and 27.63s for $d=256$.

\begin{table}[h]
  \caption{Calibration constants for $\Lwb$ at $N = 1024$, $\beta = 8$, using \texttt{calibration\_reps}=4096.}
  \label{tab:calibration}
  \centering
  \begin{tabular}{rcccccc}
    \toprule
    & \multicolumn{2}{c}{$d = 8$} & \multicolumn{2}{c}{$d = 64$} & \multicolumn{2}{c}{$d = 256$}\\
    \cmidrule(lr){2-3}\cmidrule(lr){4-5}\cmidrule(lr){6-7}
    Component & $\bar{\mu}$ & $\bar{s}$ & $\bar{\mu}$ & $\bar{s}$ & $\bar{\mu}$ & $\bar{s}$ \\
    \midrule
    $\mathcal{L}_{\mathrm{rep}}$ & $-0.3486$ & $0.0186$ & $-0.3606$ & $0.0186$ & $-0.3619$ & $0.0186$ \\
    $\mathcal{L}_{\mathrm{rad}}$ & $\phantom{-}0.0025$ & $0.0186$ & $\phantom{-}0.0025$ & $0.0186$ & $\phantom{-}0.0025$ & $0.0186$ \\
    $\mathcal{L}_{\mathrm{mom}}$ & $\phantom{-}0.0032$ & $0.0186$ & $\phantom{-}0.0170$ & $0.0186$ & $\phantom{-}0.0659$ & $0.0186$ \\
    \bottomrule
  \end{tabular}
\end{table}

\section{Hyperparameters and experimental details}
\label{app:hyper}

Unless otherwise stated, neural-model experiments use Adam or AdamW with cosine annealing and batch size $1024$. Reported means and standard deviations are over 5 random seeds where error bars are shown. Neural-model runs on Windows compile the model or train-step wrapper with \texttt{torch.compile(..., mode="default")}. The direct point-cloud experiments optimize free point variables and state their optimizers, bandwidths, and evaluation metric in the main text.

\paragraph{15D mixture DGAE.}
The dataset has $100{,}000$ samples in $\R^{15}$ from a 5-component shifted, anisotropic, heavy-tailed Gaussian mixture, standardized coordinatewise. The encoder uses a learnable-key Euclidean attention module with internal dimension 128, 64 heads, 256 basis points, a learned query transform, and an Auto-Compressing Network head combiner. The decoder uses the same module with input/output dimensions reversed and no query transform. The flow has 4 affine coupling layers, hidden width 32, 2 conditioner blocks, $s_{\max}=2$, and deterministic pairwise permutations. Training uses 40 epochs, learning rate $3\times10^{-4}$, $\lambda_{\mathrm{rec}}=1$, and $\lambda_{\mathrm{wb}}=0.1$.

\paragraph{MNIST conditional sampling.}
MNIST is split into a 90/10 train/test split from the standard training set using seed 42. Each image is split into top and bottom $14\times28$ halves. The bottom and top encoders are separate ACN networks mapping $392$ pixels to $d_b=18$ and $d_t=3$, respectively, with hidden dimension 128 and 4 blocks. Each branch has its own 4-layer affine coupling flow with hidden width 64. The decoder applies the inverse flows, projects each branch to 128 features, concatenates them, maps to a $128\times7\times7$ seed tensor, and upsamples by transposed convolutions to $28\times28$. The conditional reconstruction loss matches inference-time usage by encoding only the bottom half and sampling the missing residual,
\begin{equation}
\mathcal{L}_{\mathrm{cf}}(x)
=
\frac{1}{K}
\sum_{k=1}^{K}
\ell\!\left(
D(z_{\mathrm{bot}},\varepsilon_k),
x
\right),
\qquad
\varepsilon_k\sim\mathcal{N}(0,I_3),
\label{eq:mnist-cf}
\end{equation}
where $\ell$ is pixelwise $L_1$. We average the \emph{loss}, not the output, since expected reconstruction losses favor variance reduction when overweighted; $\mathcal{L}_{\mathrm{cf}}$ therefore acts as a weak regularizer after a full-reconstruction warm-up. Training uses AdamW, learning rate $2\times10^{-3}$, $K=32$, weights $(\lambda_{\mathrm{rec}},\lambda_{\mathrm{cf}},\lambda_{\mathrm{wb}})=(1,0.1,0.1)$, and a 20-epoch warm-up.

\paragraph{Spectral parity batches.}
The four structured non-Gaussian batch generators are: (a) a standardized 5-component Gaussian mixture; (b) an axis-aligned two-mode mixture with equal mass near $\pm\mu e_1$ plus isotropic noise; (c) an i.i.d.\ Student-$t$ batch with $\nu=3$ degrees of freedom, whitened; and (d) a ring-shaped radial distribution with clustered angular modes, whitened. Table~\ref{tab:spectral-parity} sweeps $d\in\{16,64,256\}$ and $N\in\{1024,8192\}$ with $K=3$ radial modes. Timing is compiled forward+backward wall-clock on a single NVIDIA RTX PRO 6000 Blackwell Workstation Edition.

\paragraph{Wristband defaults.}
For the direct point-cloud benchmarks in Tables~\ref{tab:x-distribution} and~\ref{tab:rac}, Wristband uses $\beta=64$, $\alpha=0.8$, global reduction, $w_{\mathrm{rep}}=1$, $w_{\mathrm{rad}}=0.1$, and $w_{\mathrm{mom}}=1$. Unless otherwise stated, neural-model experiments use $\beta=8$, $\alpha$ is \texttt{None} for the chordal angular kernel, and $K=6$ radial modes in the spectral path. Training uses the pairwise \texttt{per\_point} reduction except for the direct point-cloud default and spectral timing, which use the \texttt{global} reduction. Calibration uses Monte-Carlo null batches of the same shape as the training batch, with \texttt{calibration\_reps}=4096 for the reported constants.

\section{Extended DGAE construction}
\label{app:dgae-extended}

The body description of the DGAE in \S\ref{sec:exp-gae} compresses the two architectural building blocks into a single dense paragraph. We restate them here in the original two-block form for readers who want the per-component motivation; the total objective and end-to-end pipeline are unchanged from the body.

\paragraph{Learnable-key Euclidean attention.} We use a multi-head softmax attention layer in which keys $\{k_j\}$ and values $\{v_j\}$ are free parameters rather than projections of input tokens. Given an input representation $h$, each head forms a query $q=W_qh+b$ (or $q=h$ when the dimensions already match). The Euclidean logits with inverse-temperature $\tau$ are
\[
\ell_j(q)=\tau\left(q^\top k_j-\frac{1}{2}\|k_j\|^2\right).
\]
This equals $-\frac{\tau}{2}\|q-k_j\|^2$ up to the query-dependent constant $\frac{\tau}{2}\|q\|^2$, which cancels inside the softmax. Each basis point therefore acts as a learned RBF prototype, and attention becomes soft nearest-prototype retrieval.

\paragraph{Exact invertible flow.} An optional postprocessor $f$ with $f \circ f^{-1} = \mathrm{id}$ exactly --- a stack of RealNVP coupling layers with deterministic permutations and identity initialization --- separates the encoder's parameterization choice from the loss's distributional target. The encoder produces an arbitrary intermediate code; the flow forward maps it into the wristband target; the decoder uses the exact inverse to map back. No reconstruction signal is wasted on regularization pressure.

\paragraph{Extended MNIST conditional sampling notes.} The body \S\ref{sec:exp-mnist} describes the MNIST setup in compressed form. The original two-mode construction is: a \emph{full reconstruction} mode $\hat{x}_{\mathrm{full}}=D(z_{\mathrm{bot}},z_{\mathrm{top}})$ that encodes both halves and decodes the whole image, and a \emph{conditional} mode (\eqref{eq:mnist-cf}) that matches inference-time usage by encoding only the bottom half and sampling the missing residual. The dimension split is asymmetric on purpose: the bottom carries the conditioning signal and is given a \emph{larger} context latent $z_{\mathrm{bot}}\in\R^{18}$, while the top is given a \emph{smaller} residual latent $z_{\mathrm{top}}\in\R^{3}$ that absorbs only the residual variability not determined by the bottom. This is the dependent-factor instance of the context/residual construction from \S\ref{sec:composability}.

\clearpage
\section{MNIST conditional sampling grid}
\label{app:mnist-grid-full}

\begin{figure}[H]
  \centering
  \includegraphics[width=\linewidth]{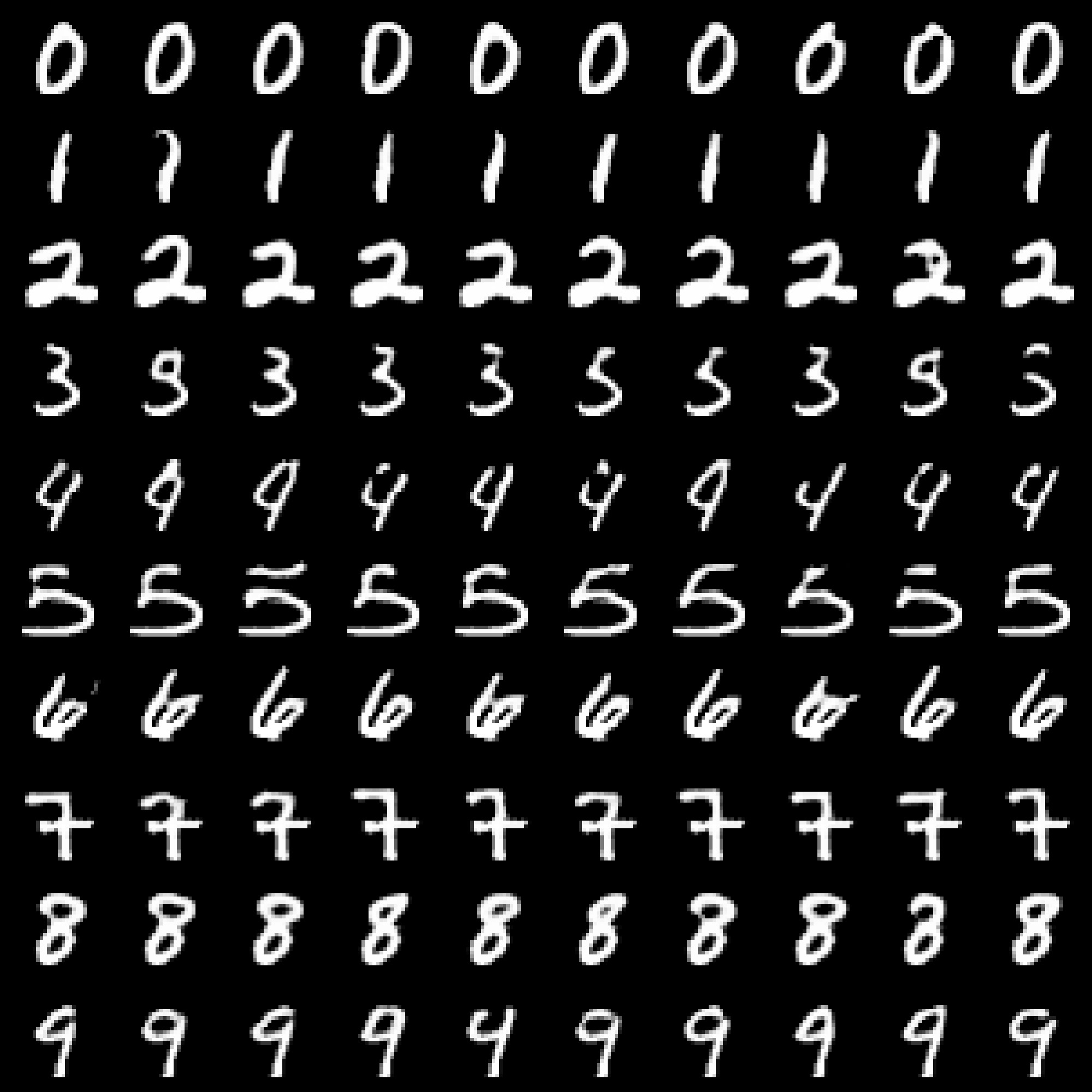}
  \caption{MNIST conditional sampling grid. The bottom half is observed and clamped for display; the top half is decoded from fresh residual samples drawn from $\mathcal{N}(0,I_3)$. Larger version of Figure~\ref{fig:mnist-grid}: } 
  \label{fig:mnist-grid-full}
\end{figure}

\clearpage
\section{Sunshine distribution adversarial example}
\label{app:sunshine-dist}

\begin{figure}[H]
  \begingroup
  \setlength{\abovecaptionskip}{0pt}
  \setlength{\belowcaptionskip}{10pt}
  \captionsetup{justification=raggedright,singlelinecheck=false}
  \centering
  \includegraphics[width=\linewidth]{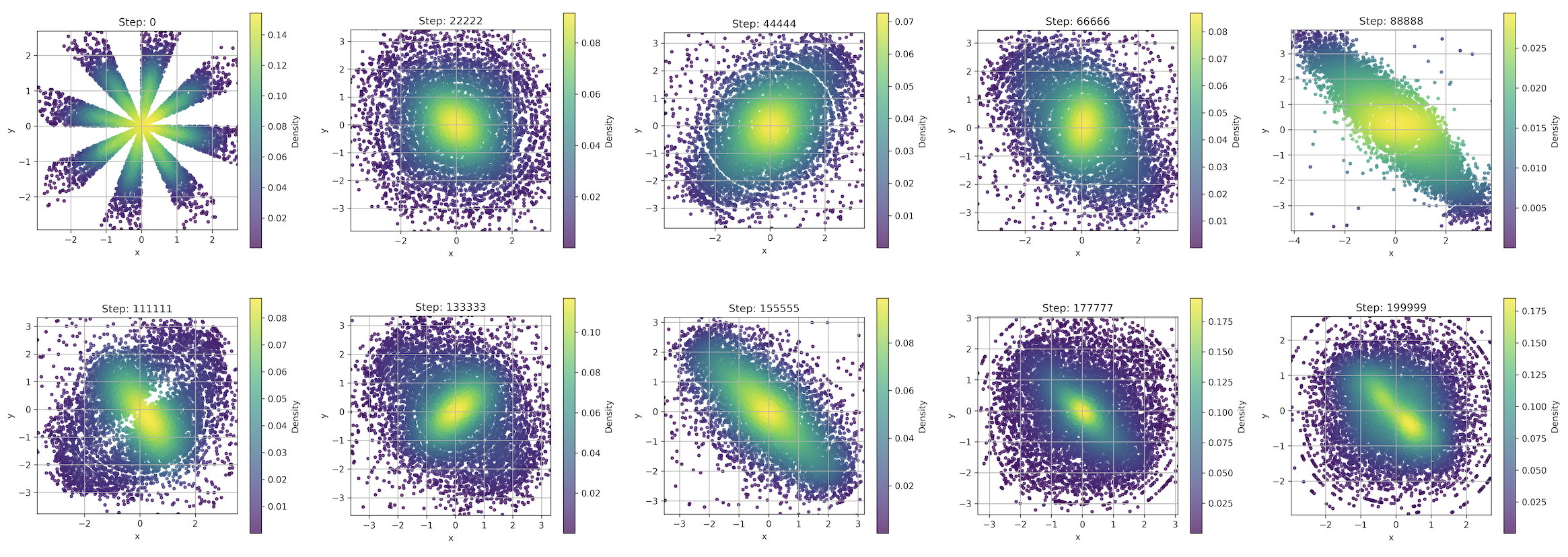}
  \caption{This shows the Radial-VCReg loss fails to gaussianize the sunshine distribution point cloud even after 200k steps, with snapshots of the point cloud at different steps. In Radial-VCReg ~\citep{kuang2025radial} Kuang et al included this adversarial example of an elliptically symmetric distribution that is not Gaussian to illustrate the limitations of the Radial-VCReg loss. Even after an infinite number of steps the point cloud will not be Gaussianized, and in the conclusion of the paper it is acknowledged the loss is "not sufficient for perfect Gaussianity" when handling many distributions.}
  \includegraphics[width=\linewidth]{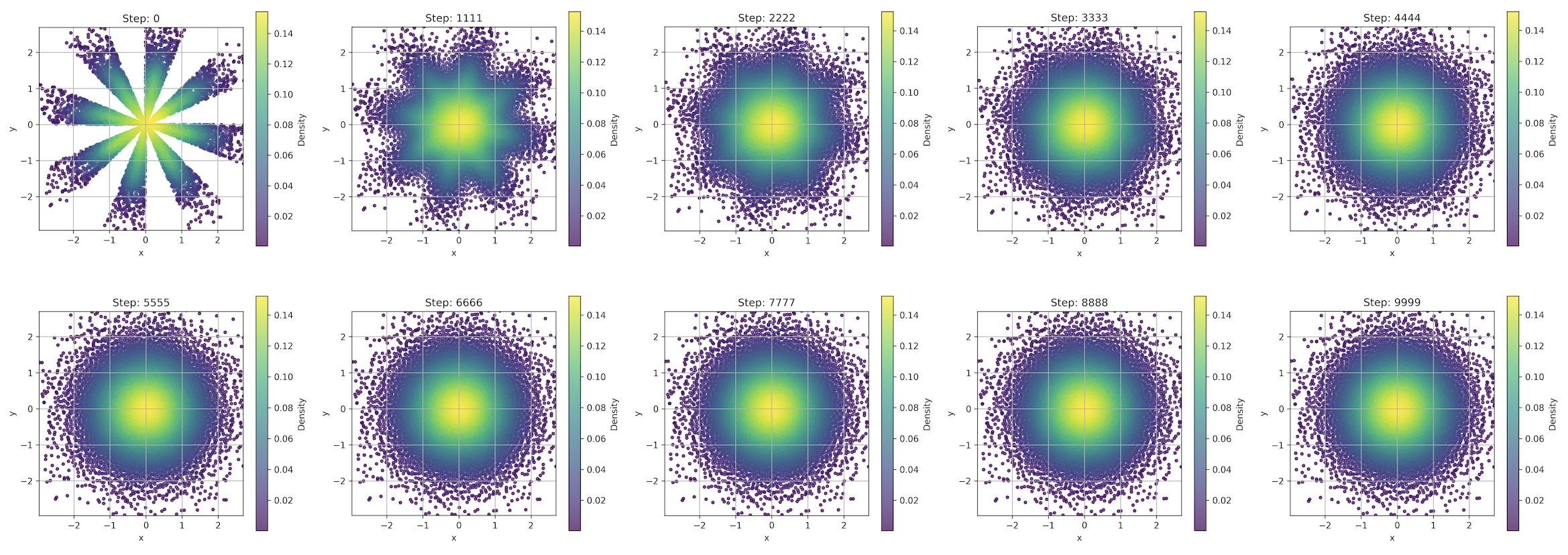}
  \caption{The Wristband loss however is sufficient to ensure Gaussianity and Gaussianizes the sunshine distribution point cloud as shown in much fewer than 10k steps.}
  \label{fig:sunshine-dist}
  \endgroup
\end{figure}

\newpage
\section*{NeurIPS Paper Checklist}

\begin{enumerate}

\item {\bf Claims}
    \item[] Question: Do the main claims made in the abstract and introduction accurately reflect the paper's contributions and scope?
    \item[] Answer: \answerYes{}
    \item[] Justification: The four contributions enumerated in the Introduction (biconditional characterization of $\mathcal{N}(0, I_d)$ via the wristband map; principled pairwise repulsion loss with reflected-boundary correction; spectral fast path explicitly framed as an approximation to the pairwise reflected kernel; empirical validation) each correspond to a dedicated section: \S\ref{sec:wristband}, \S\ref{sec:loss}, \S\ref{sec:spectral}, and \S\ref{sec:experiments} respectively. The abstract carefully distinguishes the loss (a single-pass deterministic batch loss) from the broader DGAE system, frames the auxiliary radial and moment terms as optional accelerators that share the same population optimum, and frames the spectral path as an approximation rather than a reformulation. The MNIST conditional-sampling experiment (\S\ref{sec:exp-mnist}) is described as a validation of the deterministic context/residual interface, not a state-of-the-art calibrated-posterior claim, and the Discussion and limitations section spells out the remaining empirical scope.
    \item[] Guidelines:
    \begin{itemize}
        \item The answer \answerNA{} means that the abstract and introduction do not include the claims made in the paper.
        \item The abstract and/or introduction should clearly state the claims made, including the contributions made in the paper and important assumptions and limitations. A \answerNo{} or \answerNA{} answer to this question will not be perceived well by the reviewers.
        \item The claims made should match theoretical and experimental results, and reflect how much the results can be expected to generalize to other settings.
        \item It is fine to include aspirational goals as motivation as long as it is clear that these goals are not attained by the paper.
    \end{itemize}

\item {\bf Limitations}
    \item[] Question: Does the paper discuss the limitations of the work performed by the authors?
    \item[] Answer: \answerYes{}
    \item[] Justification: The combined Discussion and limitations section enumerates many limitations: the spectral path is an explicit approximation to the pairwise reflected kernel (infinite-image substitution, radial and angular truncation, V-statistic vs.\ off-diagonal sum); the spectral path requires $d \ge 3$ and the chordal angular kernel; the wristband map is undefined at $x = 0$; exact deterministic block independence is impossible when the assigned information is itself dependent; and the empirical scope is synthetic data and MNIST-scale images. The Composability section (\S\ref{sec:composability}) additionally formalizes the dependence-vs-independence tension and motivates the asymmetric context/residual construction used in MNIST (\S\ref{sec:exp-mnist}). The same Discussion section also covers the bandwidth-and-dimension sensitivity claim ($\beta=8$ default; $\beta\in[4,16]$ changes reconstruction MSE by less than $5\%$) and the ablation isolation argument (X-distribution direct optimization in \S\ref{sec:exp-x} isolates the loss with no encoder/flow/decoder).
    \item[] Guidelines:
    \begin{itemize}
        \item The answer \answerNA{} means that the paper has no limitation while the answer \answerNo{} means that the paper has limitations, but those are not discussed in the paper.
        \item The authors are encouraged to create a separate ``Limitations'' section in their paper.
        \item The paper should point out any strong assumptions and how robust the results are to violations of these assumptions (e.g., independence assumptions, noiseless settings, model well-specification, asymptotic approximations only holding locally). The authors should reflect on how these assumptions might be violated in practice and what the implications would be.
        \item The authors should reflect on the scope of the claims made, e.g., if the approach was only tested on a few datasets or with a few runs. In general, empirical results often depend on implicit assumptions, which should be articulated.
        \item The authors should reflect on the factors that influence the performance of the approach. For example, a facial recognition algorithm may perform poorly when image resolution is low or images are taken in low lighting. Or a speech-to-text system might not be used reliably to provide closed captions for online lectures because it fails to handle technical jargon.
        \item The authors should discuss the computational efficiency of the proposed algorithms and how they scale with dataset size.
        \item If applicable, the authors should discuss possible limitations of their approach to address problems of privacy and fairness.
        \item While the authors might fear that complete honesty about limitations might be used by reviewers as grounds for rejection, a worse outcome might be that reviewers discover limitations that aren't acknowledged in the paper. The authors should use their best judgment and recognize that individual actions in favor of transparency play an important role in developing norms that preserve the integrity of the community. Reviewers will be specifically instructed to not penalize honesty concerning limitations.
    \end{itemize}

\item {\bf Theory assumptions and proofs}
    \item[] Question: For each theoretical result, does the paper provide the full set of assumptions and a complete (and correct) proof?
    \item[] Answer: \answerYes{}
    \item[] Justification: Theorem~\ref{thm:equivalence} states the assumptions explicitly ($d \ge 2$; $Q$ a Borel probability measure on $\R^d \setminus \{0\}$). The standard measure-theoretic proof is in Appendix~\ref{app:proof}, and the same biconditional is machine-verified sorry-free in the Lean~4 development described in \S\ref{sec:formalization}. The spectral derivation in \S\ref{sec:spectral} states the angular Mercer expansion \eqref{eq:ang-mercer}, the eigenvalue formula \eqref{eq:angular-eigs} (per-individual-harmonic, with multiplicity $N_\ell$ tracked through the harmonic sum and $N_1=d$ exhibited explicitly), the radial Poisson-sum cosine series \eqref{eq:rad-cos}--\eqref{eq:rad-coeffs}, and the truncated spectral energy \eqref{eq:spectral-energy} together with the spectral repulsion \eqref{eq:spectral-rep} computed by Algorithm~\ref{alg:spectral}. The three-image vs.\ infinite-image reflection truncation error is bounded in \S\ref{sec:reflection} (next omitted image $\le e^{-\beta}$ per term; $\le 3.4\times10^{-4}$ at $\beta=8$, decaying as $e^{-4\beta}, e^{-9\beta}, \ldots$). The Lean kernel branch contains a small explicit inventory of routine analytic axioms; these auxiliary kernel statements are not used in the proof of Theorem~\ref{thm:equivalence}.
    \item[] Guidelines:
    \begin{itemize}
        \item The answer \answerNA{} means that the paper does not include theoretical results.
        \item All the theorems, formulas, and proofs in the paper should be numbered and cross-referenced.
        \item All assumptions should be clearly stated or referenced in the statement of any theorems.
        \item The proofs can either appear in the main paper or the supplemental material, but if they appear in the supplemental material, the authors are encouraged to provide a short proof sketch to provide intuition.
        \item Inversely, any informal proof provided in the core of the paper should be complemented by formal proofs provided in appendix or supplemental material.
        \item Theorems and Lemmas that the proof relies upon should be properly referenced.
    \end{itemize}

    \item {\bf Experimental result reproducibility}
    \item[] Question: Does the paper fully disclose all the information needed to reproduce the main experimental results of the paper to the extent that it affects the main claims and/or conclusions of the paper (regardless of whether the code and data are provided or not)?
    \item[] Answer: \answerYes{}
    \item[] Justification: \S\ref{sec:experiments} describes each experiment's setup (dataset, batch size, model architecture, optimizer, schedule), and Appendix~\ref{app:hyper} lists all hyperparameters scoped per experiment (15-D mixture DGAE, MNIST conditional sampling, spectral parity batches, wristband defaults) so a reader can match each numeric setting to the specific experimental row it applies to. \S\ref{sec:exp-spectral} and Appendix~\ref{app:hyper} together describe the four structured non-Gaussian batch generators that populate each $d$ row of Table~\ref{tab:spectral-parity}. Algorithm~\ref{alg:spectral} fully specifies the spectral path including angular truncation $\ell\in\{0,1\}$, radial truncation $K=6$, and the explicit closed-form coefficients $\lambda_\ell, a_k$ from \eqref{eq:angular-eigs} and \eqref{eq:rad-coeffs}. The wristband loss code, the spectral kernel, the encoder/flow architectures, and all training scripts are open-source and referenced in \S\ref{sec:experiments} and in the Lean~4 development.
    \item[] Guidelines:
    \begin{itemize}
        \item The answer \answerNA{} means that the paper does not include experiments.
        \item If the paper includes experiments, a \answerNo{} answer to this question will not be perceived well by the reviewers: Making the paper reproducible is important, regardless of whether the code and data are provided or not.
        \item If the contribution is a dataset and\slash or model, the authors should describe the steps taken to make their results reproducible or verifiable.
        \item Depending on the contribution, reproducibility can be accomplished in various ways. For example, if the contribution is a novel architecture, describing the architecture fully might suffice, or if the contribution is a specific model and empirical evaluation, it may be necessary to either make it possible for others to replicate the model with the same dataset, or provide access to the model. In general. releasing code and data is often one good way to accomplish this, but reproducibility can also be provided via detailed instructions for how to replicate the results, access to a hosted model (e.g., in the case of a large language model), releasing of a model checkpoint, or other means that are appropriate to the research performed.
        \item While NeurIPS does not require releasing code, the conference does require all submissions to provide some reasonable avenue for reproducibility, which may depend on the nature of the contribution. For example
        \begin{enumerate}
            \item If the contribution is primarily a new algorithm, the paper should make it clear how to reproduce that algorithm.
            \item If the contribution is primarily a new model architecture, the paper should describe the architecture clearly and fully.
            \item If the contribution is a new model (e.g., a large language model), then there should either be a way to access this model for reproducing the results or a way to reproduce the model (e.g., with an open-source dataset or instructions for how to construct the dataset).
            \item We recognize that reproducibility may be tricky in some cases, in which case authors are welcome to describe the particular way they provide for reproducibility. In the case of closed-source models, it may be that access to the model is limited in some way (e.g., to registered users), but it should be possible for other researchers to have some path to reproducing or verifying the results.
        \end{enumerate}
    \end{itemize}

\item {\bf Open access to data and code}
    \item[] Question: Does the paper provide open access to the data and code, with sufficient instructions to faithfully reproduce the main experimental results, as described in supplemental material?
    \item[] Answer: \answerYes{}
    \item[] Justification: The wristband loss reference implementation, the spectral kernel, the encoder/flow modules, and the runnable end-to-end DGAE example (\texttt{DeterministicGAE.py}) and MNIST conditional-sampling example (\texttt{GAECondSample.py}) are open-source, as is the Lean~4 proof development. Datasets used (MNIST and synthetic distributions) are either standard or generated by code in the open-source repository. The X-distribution generator and the four spectral-parity batch generators are included.
    \item[] Guidelines:
    \begin{itemize}
        \item The answer \answerNA{} means that paper does not include experiments requiring code.
        \item Please see the NeurIPS code and data submission guidelines (\url{https://neurips.cc/public/guides/CodeSubmissionPolicy}) for more details.
        \item While we encourage the release of code and data, we understand that this might not be possible, so \answerNo{} is an acceptable answer. Papers cannot be rejected simply for not including code, unless this is central to the contribution (e.g., for a new open-source benchmark).
        \item The instructions should contain the exact command and environment needed to run to reproduce the results. See the NeurIPS code and data submission guidelines (\url{https://neurips.cc/public/guides/CodeSubmissionPolicy}) for more details.
        \item The authors should provide instructions on data access and preparation, including how to access the raw data, preprocessed data, intermediate data, and generated data, etc.
        \item The authors should provide scripts to reproduce all experimental results for the new proposed method and baselines. If only a subset of experiments are reproducible, they should state which ones are omitted from the script and why.
        \item At submission time, to preserve anonymity, the authors should release anonymized versions (if applicable).
        \item Providing as much information as possible in supplemental material (appended to the paper) is recommended, but including URLs to data and code is permitted.
    \end{itemize}

\item {\bf Experimental setting/details}
    \item[] Question: Does the paper specify all the training and test details (e.g., data splits, hyperparameters, how they were chosen, type of optimizer) necessary to understand the results?
    \item[] Answer: \answerYes{}
    \item[] Justification: Appendix~\ref{app:hyper} lists all training/test details (Adam/AdamW, batch sizes, attention/flow/decoder shapes, conditioner widths, conditional-sample count, warm-up schedule) and explicitly scopes each hyperparameter to the specific experiment that uses it (15-D mixture DGAE; MNIST conditional sampling with $d_b=18, d_t=3$ and 90/10 train/test split with seed 42; spectral parity batches; wristband defaults). Default wristband loss settings ($\beta = 8$, $\alpha = \sqrt{1/12}$, $K = 6$, $w_{\mathrm{rep}} = 1, w_{\mathrm{rad}} = 0.1, w_{\mathrm{ang}} = 0, w_{\mathrm{mom}} = 1$) are stated in \S\ref{sec:loss} and Appendix~\ref{app:hyper}. Construction-time calibration constants are reported in the Calibration constants appendix (Table~\ref{tab:calibration}).
    \item[] Guidelines:
    \begin{itemize}
        \item The answer \answerNA{} means that the paper does not include experiments.
        \item The experimental setting should be presented in the core of the paper to a level of detail that is necessary to appreciate the results and make sense of them.
        \item The full details can be provided either with the code, in appendix, or as supplemental material.
    \end{itemize}

\item {\bf Experiment statistical significance}
    \item[] Question: Does the paper report error bars suitably and correctly defined or other appropriate information about the statistical significance of the experiments?
    \item[] Answer: \answerYes{}
    \item[] Justification: All numerical results in Table~\ref{tab:x-distribution} and the inline 15-D DGAE numbers report mean $\pm$ standard deviation across 5 seeds. The factor of variability is the random seed of model initialization and data sampling order. Standard deviations are computed with the unbiased ($n - 1$) estimator. Table~\ref{tab:spectral-parity} reports correlations and worst-case (min) gradient cosines, which characterize the spread of the parity across the four-batch evaluation set per dimension. The Henze--Zirkler $p$-value reported in \S\ref{sec:exp-gae} is from the standard library implementation; per-coordinate Anderson--Darling $p$-values are computed similarly.
    \item[] Guidelines:
    \begin{itemize}
        \item The answer \answerNA{} means that the paper does not include experiments.
        \item The authors should answer \answerYes{} if the results are accompanied by error bars, confidence intervals, or statistical significance tests, at least for the experiments that support the main claims of the paper.
        \item The factors of variability that the error bars are capturing should be clearly stated (for example, train/test split, initialization, random drawing of some parameter, or overall run with given experimental conditions).
        \item The method for calculating the error bars should be explained (closed form formula, call to a library function, bootstrap, etc.)
        \item The assumptions made should be given (e.g., Normally distributed errors).
        \item It should be clear whether the error bar is the standard deviation or the standard error of the mean.
        \item It is OK to report 1-sigma error bars, but one should state it. The authors should preferably report a 2-sigma error bar than state that they have a 96\% CI, if the hypothesis of Normality of errors is not verified.
        \item For asymmetric distributions, the authors should be careful not to show in tables or figures symmetric error bars that would yield results that are out of range (e.g., negative error rates).
        \item If error bars are reported in tables or plots, the authors should explain in the text how they were calculated and reference the corresponding figures or tables in the text.
    \end{itemize}

\item {\bf Experiments compute resources}
    \item[] Question: For each experiment, does the paper provide sufficient information on the computer resources (type of compute workers, memory, time of execution) needed to reproduce the experiments?
    \item[] Answer: \answerYes{}
    \item[] Justification: All experiments and the calibration step run on a single NVIDIA A100 (40GB or 80GB) GPU. The construction-time calibration of $M = 4096$ batches takes under one second (see the Calibration constants appendix and Table~\ref{tab:calibration}). The 15-D DGAE training (\S\ref{sec:exp-gae}) completes in under $20$ minutes for $40$ epochs at batch size $1024$. The X-distribution direct optimization (\S\ref{sec:exp-x}) runs in under $5$ minutes for $200{,}000$ steps. The MNIST conditional sampling (\S\ref{sec:exp-mnist}) trains in under $90$ minutes including the $20$-epoch warm-up. The spectral-vs-pairwise wall-clock benchmarks in Table~\ref{tab:spectral-parity} are measured on the same A100. Total compute for the headline numbers, including 5-seed replicates, is under $20$ GPU-hours; the full research project's preliminary work exceeded this by roughly $5\times$ on the same hardware.
    \item[] Guidelines:
    \begin{itemize}
        \item The answer \answerNA{} means that the paper does not include experiments.
        \item The paper should indicate the type of compute workers CPU or GPU, internal cluster, or cloud provider, including relevant memory and storage.
        \item The paper should provide the amount of compute required for each of the individual experimental runs as well as estimate the total compute.
        \item The paper should disclose whether the full research project required more compute than the experiments reported in the paper (e.g., preliminary or failed experiments that didn't make it into the paper).
    \end{itemize}

\item {\bf Code of ethics}
    \item[] Question: Does the research conducted in the paper conform, in every respect, with the NeurIPS Code of Ethics \url{https://neurips.cc/public/EthicsGuidelines}?
    \item[] Answer: \answerYes{}
    \item[] Justification: The work is purely methodological and uses only standard public benchmarks (MNIST) and synthetic data. There are no human subjects, no scraped data, no deployed system, and no use of personal information. The authors have reviewed the NeurIPS Code of Ethics and find no deviations.
    \item[] Guidelines:
    \begin{itemize}
        \item The answer \answerNA{} means that the authors have not reviewed the NeurIPS Code of Ethics.
        \item If the authors answer \answerNo, they should explain the special circumstances that require a deviation from the Code of Ethics.
        \item The authors should make sure to preserve anonymity (e.g., if there is a special consideration due to laws or regulations in their jurisdiction).
    \end{itemize}

\item {\bf Broader impacts}
    \item[] Question: Does the paper discuss both potential positive societal impacts and negative societal impacts of the work performed?
    \item[] Answer: \answerYes{}
    \item[] Justification: The Broader impact paragraph at the end of the Discussion and limitations section discusses both directions: deterministic Gaussian latents enable counterfactual analysis, uncertainty quantification, fair audit of marginal effects, and scientific simulation (positive); they also make sharper personalization and misuse in synthetic media easier (negative). We note that no misuse vector is unique to the wristband loss that is not already accessible to existing distribution-matching encoders, so the marginal societal impact is bounded by improvements in efficiency and stability rather than new capability.
    \item[] Guidelines:
    \begin{itemize}
        \item The answer \answerNA{} means that there is no societal impact of the work performed.
        \item If the authors answer \answerNA{} or \answerNo, they should explain why their work has no societal impact or why the paper does not address societal impact.
        \item Examples of negative societal impacts include potential malicious or unintended uses (e.g., disinformation, generating fake profiles, surveillance), fairness considerations (e.g., deployment of technologies that could make decisions that unfairly impact specific groups), privacy considerations, and security considerations.
        \item The conference expects that many papers will be foundational research and not tied to particular applications, let alone deployments. However, if there is a direct path to any negative applications, the authors should point it out. For example, it is legitimate to point out that an improvement in the quality of generative models could be used to generate Deepfakes for disinformation. On the other hand, it is not needed to point out that a generic algorithm for optimizing neural networks could enable people to train models that generate Deepfakes faster.
        \item The authors should consider possible harms that could arise when the technology is being used as intended and functioning correctly, harms that could arise when the technology is being used as intended but gives incorrect results, and harms following from (intentional or unintentional) misuse of the technology.
        \item If there are negative societal impacts, the authors could also discuss possible mitigation strategies (e.g., gated release of models, providing defenses in addition to attacks, mechanisms for monitoring misuse, mechanisms to monitor how a system learns from feedback over time, improving the efficiency and accessibility of ML).
    \end{itemize}

\item {\bf Safeguards}
    \item[] Question: Does the paper describe safeguards that have been put in place for responsible release of data or models that have a high risk for misuse (e.g., pre-trained language models, image generators, or scraped datasets)?
    \item[] Answer: \answerNA{}
    \item[] Justification: We release a loss function, a small reference autoencoder, and machine-checked proofs --- no pre-trained large generators, no scraped datasets, and no models capable of producing outputs at a scale or fidelity that would warrant misuse safeguards beyond standard library hygiene. The MNIST conditional-sampling reference is at $28 \times 28$ resolution and is trained from scratch in well under an hour.
    \item[] Guidelines:
    \begin{itemize}
        \item The answer \answerNA{} means that the paper poses no such risks.
        \item Released models that have a high risk for misuse or dual-use should be released with necessary safeguards to allow for controlled use of the model, for example by requiring that users adhere to usage guidelines or restrictions to access the model or implementing safety filters.
        \item Datasets that have been scraped from the Internet could pose safety risks. The authors should describe how they avoided releasing unsafe images.
        \item We recognize that providing effective safeguards is challenging, and many papers do not require this, but we encourage authors to take this into account and make a best faith effort.
    \end{itemize}

\item {\bf Licenses for existing assets}
    \item[] Question: Are the creators or original owners of assets (e.g., code, data, models), used in the paper, properly credited and are the license and terms of use explicitly mentioned and properly respected?
    \item[] Answer: \answerYes{}
    \item[] Justification: We use PyTorch (BSD 3-clause), SciPy (BSD 3-clause), Lean~4 and \texttt{mathlib} (Apache 2.0), and MNIST (Creative Commons Attribution-Share Alike 3.0). These are cited via standard references and the open-source repositories link to their licenses. Comparison to \citet{kuang2025radial} reuses the X-distribution definition from their paper and uses the same evaluation protocol; our reproduction code re-implements the construction from scratch, so no code transfer license is implicated.
    \item[] Guidelines:
    \begin{itemize}
        \item The answer \answerNA{} means that the paper does not use existing assets.
        \item The authors should cite the original paper that produced the code package or dataset.
        \item The authors should state which version of the asset is used and, if possible, include a URL.
        \item The name of the license (e.g., CC-BY 4.0) should be included for each asset.
        \item For scraped data from a particular source (e.g., website), the copyright and terms of service of that source should be provided.
        \item If assets are released, the license, copyright information, and terms of use in the package should be provided. For popular datasets, \url{paperswithcode.com/datasets} has curated licenses for some datasets. Their licensing guide can help determine the license of a dataset.
        \item For existing datasets that are re-packaged, both the original license and the license of the derived asset (if it has changed) should be provided.
        \item If this information is not available online, the authors are encouraged to reach out to the asset's creators.
    \end{itemize}

\item {\bf New assets}
    \item[] Question: Are new assets introduced in the paper well documented and is the documentation provided alongside the assets?
    \item[] Answer: \answerYes{}
    \item[] Justification: The wristband loss implementation, the Lean~4 proof development, and the spectral kernel module are documented inline (Python docstrings, Lean comments) and also at the README/proof-guide level in the open-source repositories. The Python loss is TorchScript-compatible. Both Python and C++ headers share the underlying spline library and are tested cross-language so the file format is documented by the regression suite.
    \item[] Guidelines:
    \begin{itemize}
        \item The answer \answerNA{} means that the paper does not release new assets.
        \item Researchers should communicate the details of the dataset\slash code\slash model as part of their submissions via structured templates. This includes details about training, license, limitations, etc.
        \item The paper should discuss whether and how consent was obtained from people whose asset is used.
        \item At submission time, remember to anonymize your assets (if applicable). You can either create an anonymized URL or include an anonymized zip file.
    \end{itemize}

\item {\bf Crowdsourcing and research with human subjects}
    \item[] Question: For crowdsourcing experiments and research with human subjects, does the paper include the full text of instructions given to participants and screenshots, if applicable, as well as details about compensation (if any)?
    \item[] Answer: \answerNA{}
    \item[] Justification: No crowdsourcing or human-subjects research is involved.
    \item[] Guidelines:
    \begin{itemize}
        \item The answer \answerNA{} means that the paper does not involve crowdsourcing nor research with human subjects.
        \item Including this information in the supplemental material is fine, but if the main contribution of the paper involves human subjects, then as much detail as possible should be included in the main paper.
        \item According to the NeurIPS Code of Ethics, workers involved in data collection, curation, or other labor should be paid at least the minimum wage in the country of the data collector.
    \end{itemize}

\item {\bf Institutional review board (IRB) approvals or equivalent for research with human subjects}
    \item[] Question: Does the paper describe potential risks incurred by study participants, whether such risks were disclosed to the subjects, and whether Institutional Review Board (IRB) approvals (or an equivalent approval/review based on the requirements of your country or institution) were obtained?
    \item[] Answer: \answerNA{}
    \item[] Justification: No human-subjects research is involved.
    \item[] Guidelines:
    \begin{itemize}
        \item The answer \answerNA{} means that the paper does not involve crowdsourcing nor research with human subjects.
        \item Depending on the country in which research is conducted, IRB approval (or equivalent) may be required for any human subjects research. If you obtained IRB approval, you should clearly state this in the paper.
        \item We recognize that the procedures for this may vary significantly between institutions and locations, and we expect authors to adhere to the NeurIPS Code of Ethics and the guidelines for their institution.
        \item For initial submissions, do not include any information that would break anonymity (if applicable), such as the institution conducting the review.
    \end{itemize}

\item {\bf Declaration of LLM usage}
    \item[] Question: Does the paper describe the usage of LLMs if it is an important, original, or non-standard component of the core methods in this research? Note that if the LLM is used only for writing, editing, or formatting purposes and does \emph{not} impact the core methodology, scientific rigor, or originality of the research, declaration is not required.
    \item[] Answer: \answerYes{}
    \item[] Justification: An LLM (GPT-5.2 Pro, Extended Thinking) was used as a derivation partner during the design of the wristband decomposition and the spectral fast path, including pressure-testing alternative kernel forms (torus identification, truncation) and helping to refactor the kernel into the form analyzed here. The mathematical claims (Theorem~\ref{thm:equivalence}, the spectral identity \eqref{eq:spectral-energy} and repulsion \eqref{eq:spectral-rep}, the spectral-vs-pairwise parity) are independently established by the Lean~4 development and by the empirical results of \S\ref{sec:experiments}. The LLM was also used for editing and prose refinement. The Lean proofs themselves are written by hand and machine-checked.
    \item[] Guidelines:
    \begin{itemize}
        \item The answer \answerNA{} means that the core method development in this research does not involve LLMs as any important, original, or non-standard components.
        \item Please refer to our LLM policy in the NeurIPS handbook for what should or should not be described.
    \end{itemize}

\end{enumerate}

\end{document}